\documentclass[sigconf]{acmart}

\AtBeginDocument{
  }

\setcopyright{acmlicensed}
\copyrightyear{2025}
\acmYear{2025}
\acmDOI{XXXXXXX.XXXXXXX}
\acmConference[RecSys '25]{Proceedings of the 19th ACM Conference on Recommender Systems}{October 6--10, 2025}{Prague, Czech Republic}
\acmISBN{978-1-4503-XXXX-X/2025/10}

\usepackage{amsmath}

\usepackage{amssymb}
\usepackage{booktabs}
\usepackage{multirow}
\usepackage{xcolor}
\usepackage{graphicx}
\usepackage{tikz}
\usepackage{bbm}

\usepackage{pgfplots}
\pgfplotsset{compat=1.17}
\begin{document}


\title{Adaptive Loss Balancing for Noise-Robust GRPO in Generative Recommendation}

\author{Kewei Xu}
\authornote{Both authors contributed equally to this research.}
\email{xukewei3@jd.com} 
\affiliation{
    \institution{JD.com}
    \city{Beijing} 
    \country{China}
}

\author{Junbo Qi}
\authornotemark[1]
\email{junboqi@toki.waseda.jp}
\affiliation{
    \institution{Waseda University}
    \city{Tokyo}
    \country{Japan}
}

\author{Yanyan Zou}
\authornote{Project lead.} 
\email{zoe.yyzou@gmail.com} 
\affiliation{
    \institution{JD.com}
    \city{Beijing}
    \country{China}
}

\author{Pengfei Zhang}
\email{202421080437@std.uestc.edu.cn}
\affiliation{
    \institution{University of Electronic Science and Technology of China} 
    \city{Chengdu}
    \country{China}
}

\author{Xingzhi Yao}
\email{yaoxingzhi1@jd.com} 
\affiliation{
    \institution{JD.com}
    \city{Beijing}
    \country{China}
}

\author{Shengjie Li}
\authornote{Corresponding author.} 
\email{lishengjie1@jd.com}
\affiliation{
    \institution{JD.com}
    \city{Beijing}
    \country{China}
}

\begin{abstract}

Reinforcement learning (RL) presents a promising avenue for enhancing generative recommendation beyond supervised imitation, leveraging reward signals to guide policy improvement. However, its efficacy is critically contingent on the trustworthiness of the reward model for the samples it evaluates. In practice, production rankers, the widely adopted reward models,  are trained on exposure-biased logs, leading to sample-dependent inaccuracies that violate this assumption. Our stratified analysis uncovers a consistent pattern: reward guidance is most beneficial when the policy exhibits uncertainty \emph{and} the ranker can effectively discriminate the ground-truth item from rollout negatives. On other samples, the reward signal is either negligible or detrimental, highlighting the risk of uniform RL application.

To address such an issue, we introduce \textbf{AdaGRPO}, a novel framework that treats reward-guided optimization as selective admission rather than uniform pressure. Training is anchored in supervised negative log-likelihood, while the GRPO objective is gated by a binary, per-sample clip determined by two rollout diagnostics: policy-side difficulty and reward discriminability. Instances failing either diagnostic default to pure supervision, ensuring stability and mitigating the amplification of noisy gradients. 

We validate AdaGRPO on a large-scale e-commerce dataset. At the best intermediate checkpoint, it elevates HR@10 from 11.01\% to 12.18\% while constraining hallucination below 0.22\%, and maintains robustness at the final checkpoint (HR@10 11.63\%, hallucination 0.27\%), outperforming fixed NLL--GRPO mixtures across the retrieval--validity frontier. In production A/B tests, AdaGRPO achieves statistically significant gains in click-through rate and dwell time, confirming its practical utility. These results suggest that the central challenge in applying RL to generative recommendation lies not in designing stronger rewards, but in discerning when the reward signal can be trusted to improve policy learning.

\end{abstract}



\maketitle

\section{Introduction}
\label{sec:intro}

Industrial recommendation systems are built as a cascade: a recall stage retrieves thousands of candidates from a catalogue of billions, and a ranking stage rescores them with richer features and heavier models before presentation. The recall stage prizes \emph{coverage}---it must sweep a vast item space efficiently, tolerating imprecision in exchange for not missing relevant items. The ranking stage prizes \emph{discrimination}---it fuses dense cross-features (user--item interactions, real-time context, multi-objective labels for click, conversion, and dwell time) to separate good candidates from great ones within the much smaller set it receives.

Generative retrieval (GR) reformulates the conventional recall mechanism with autoregressive decoding over learned item identifiers~\cite{geng2022recommendation,li2023text,lin2024rella}. Instead of maintaining per-item embeddings and performing approximate nearest-neighbor search, a GR model compresses the catalogue into a shared codebook of Semantic IDs and generates candidates as token sequences conditioned on the user's interaction history~\cite{NEURIPS2023_20dcab0f,onerec2024,zou2026genrecpreferenceorientedgenerativeframework}. The formulation inherits the generalization strength of large language models: items sharing semantic prefixes are decoded through common pathways, giving GR natural coverage over semantically related and long-tail products that embedding-based recall might miss. The same weight-sharing, however, limits fine-grained discrimination among items within a cluster---a capacity that ranking models acquire through dedicated per-item parameters and dense multi-objective supervision.

A natural idea is to close this precision gap with reinforcement learning: treat a multi-objective ranking model as a reward model (RM) and fine-tune the GR policy to generate candidates the RM scores favorably. Group Relative Policy Optimization (GRPO)~\cite{sheng2024grpo,deepseek2025r1}, which normalizes rewards within each rollout group and avoids a learned value network, has become the standard vehicle. Recent work on RL-based alignment for generative recommendation~\cite{onerec2024,zou2026genrecpreferenceorientedgenerativeframework} confirms that reward-guided fine-tuning can push performance beyond pure supervised imitation.

The RL recipe works cleanly in domains such as mathematical reasoning~\cite{deepseek2025r1}, where a deterministic verifier certifies each rollout as correct or incorrect. Recommendation admits no such verifier. The reward signal traces back to user feedback---clicks and purchases---that is both \emph{exposure-biased} (users interact only with items they have been shown) and \emph{sparse} (the vast majority of user--item pairs are unobserved)~\cite{joachims2017unbiased,chen2023bias}. An RM trained on these logs inherits their blind spots: it may be well-calibrated on frequently exposed items yet unreliable on long-tail or recently added products that lie outside its effective training distribution. Existing RL-for-recommendation methods, however, have not examined the resulting \emph{capability conflict} between the GR policy and the RM---a conflict rooted in their different architectures, objectives, and feature regimes---and instead apply the reward signal uniformly across all training instances.

\begin{figure}[h]
    \centering
    \includegraphics[width=\linewidth]{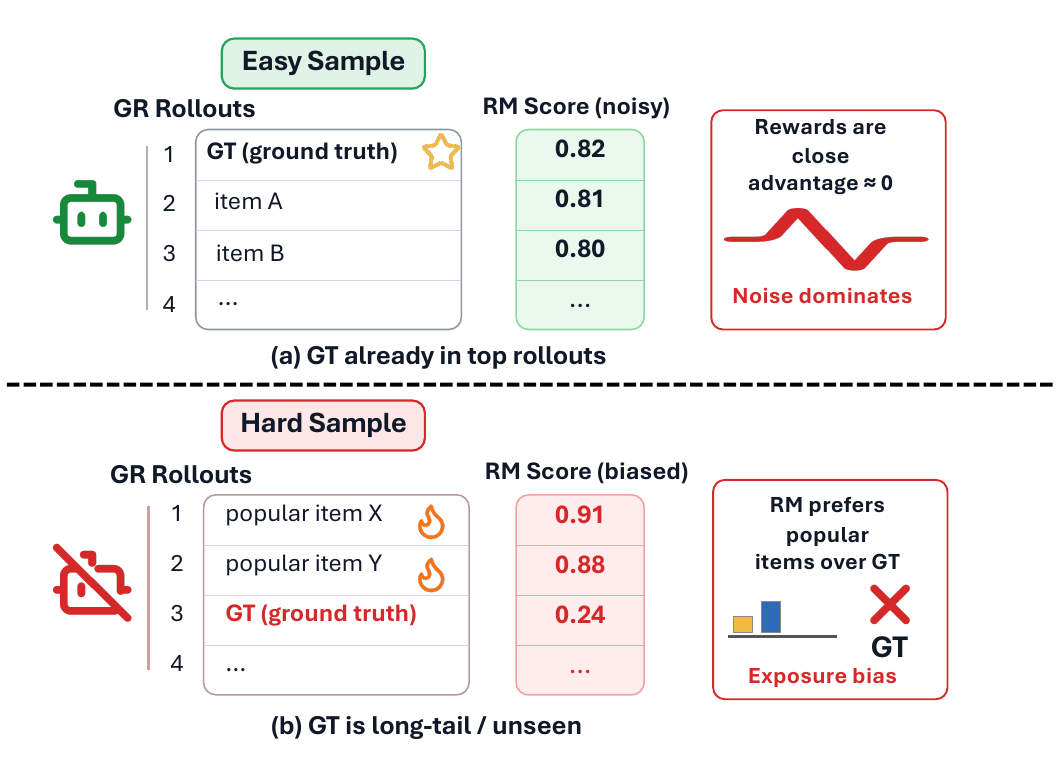}
    \caption{Two failure modes of uniform reward application. (a)~On easy samples the ground truth already ranks among top rollouts; rewards cluster tightly and the GRPO advantage is dominated by RM noise. (b)~On hard samples where the RM is unreliable, exposure bias causes it to score popular distractors above the ground truth, pushing the policy in the wrong direction.}
    \label{fig:intro}
\end{figure}

Uniform trust in the RM exposes two failure modes that pull in opposite directions (Figure~\ref{fig:intro}). On \emph{easy} instances---those where the policy already places the ground-truth item near the top of its rollout ranking---reward scores cluster tightly around the group mean. The GRPO advantage collapses to near zero and residual RM noise dominates the gradient, driving the kind of overoptimization documented in the reward-hacking literature~\cite{gao2023scaling,casper2023open}. At the other extreme, not every \emph{hard} instance benefits from RL. Some are hard precisely because the ground truth lies in a region where the RM is locally miscalibrated---long-tail products that are underrepresented in logged exposures. On these instances the RM scores popular distractors above the true target, and the resulting policy-gradient update pushes the model away from the correct answer.

A stratified analysis (Section~\ref{sec:analysis}) makes this picture precise. Aggregated over all training instances, the RM's influence on the ground-truth rank is near zero ($K{=}50$). The aggregate, however, masks a strong compositional effect: conditioned on the policy being uncertain about an instance \emph{and} the RM exhibiting clean discriminability on it, the per-instance influence becomes substantial. The RM is not globally uninformative; it is \emph{conditionally} valuable---and both conditions are computable at training time from quantities already materialized during GRPO rollout generation.

These observations motivate \textbf{AdaGRPO}, which reframes reward-guided optimization as \emph{selective admission} rather than uniform pressure. Supervised NLL anchors every training instance; the GRPO term is gated by a per-sample binary clip computed from two rank-based diagnostics on the rollout group. The first diagnostic certifies \emph{policy-side difficulty}---it fires when the ground truth falls outside the policy's top-ranked rollouts, signaling that the supervised loss alone under-specifies the correct target. The second certifies \emph{reward-side discriminability}---it requires the RM to rank the ground truth near the top while pushing contextually irrelevant negatives to the bottom, ensuring clean separation. Their conjunction admits the GRPO loss only where the policy needs help \emph{and} the RM can reliably provide it; everywhere else, the update reduces to pure supervision. Conceptually, AdaGRPO lifts the clipping principle of PPO~\cite{schulman2017proximal} from the ratio domain to the sample domain: it defines a trust region not over how far each token-level update may move, but over \emph{which instances} are permitted to contribute a policy-gradient signal at all.

We validate AdaGRPO on a large-scale e-commerce dataset. Offline, it raises HR@10 from 11.01\% to 12.18\% at the best intermediate checkpoint while holding hallucination at $\leq$0.22\%, and remains the strongest final-checkpoint model across the retrieval--validity frontier. In a production A/B test it delivers a statistically significant +0.43\% lift in effective item page views together with gains in click-through rate and dwell time. The results point to a conclusion that may generalize beyond this setting: the central challenge of RL for generative recommendation is not designing stronger rewards, but knowing when to trust them.
\section{Related Work}
\label{sec:related}

\paragraph{Generative retrieval for recommendation.}
Generative retrieval casts recommendation as autoregressive decoding over item identifiers~\cite{geng2022recommendation,onerec2024,zou2026genrecpreferenceorientedgenerativeframework}, bypassing the traditional recall--rank cascade. SFT-based alignment~\cite{tallrec2023} and retrieval-augmented variants~\cite{lin2024rella,li2023text,zheng2024harnessing} have established strong baselines; our work targets the subsequent RL fine-tuning stage.

\paragraph{RL alignment for LLMs.}
RLHF~\cite{ouyang2022training} and its derivatives---PPO~\cite{schulman2017proximal}, DPO~\cite{rafailov2023direct}, GRPO~\cite{sheng2024grpo}---form the standard toolkit, with DeepSeek-R1~\cite{deepseek2025r1} confirming that pure RL can unlock capabilities beyond SFT. PPO is the closest mechanistic precedent: it bounds each update by clipping the importance-sampling ratio (a per-token trust region). AdaGRPO clips in the sample domain, gating whether an instance contributes a policy gradient at all. In recommendation, Rank-GRPO~\cite{rankgrpo2025} introduces reward masking and ranking-importance weights, and MiniRec~\cite{minirec2025} improves sample efficiency via RL-specific filtering. These works refine \emph{how} the RM signal is used; we ask \emph{when} it should be trusted.

\paragraph{Difficulty-aware training.}
Ji et al.~\cite{ji2025difficulty} show that only intermediate-difficulty samples yield clean policy gradients in reasoning, and \cite{hardexamples2025} quantifies up to 47\% additional GRPO gain on hard instances. Methods like GRPO-LEAD~\cite{grpolead2025}, DiPO~\cite{dipo2025}, and DART-Math~\cite{dartmath2025} operationalize this through advantage reweighting and difficulty-aware rejection tuning. Translating to recommendation is non-trivial for two reasons. First, difficulty is ill-posed: user interests are implicit and contextual, so difficulty must be inferred from the policy's uncertainty over noisy behavioral logs rather than from a deterministic verifier. Second, prior work generally assumes the reward model remains trustworthy on upweighted samples. Production rankers suffer from domain-specific inaccuracies: real-time dynamics (shifting user intent, daily catalogue updates) and selection bias in logged exposures~\cite{joachims2017unbiased, chen2023bias}. Because difficulty and reward reliability fail independently, simply upweighting hard samples is unsafe. AdaGRPO addresses this by conjoining a difficulty condition with a local RM-discriminability check.
\section{Preliminaries}
\label{sec:prelim}

\subsection{Generative Recommendation as Autoregressive Decoding}
\label{sec:prelim:setup}

Let $\mathcal{V}$ denote the item catalogue. A user's interaction history $\mathbf{x} = (x_1, \ldots, x_T) \in \mathcal{V}^T$ serves as a prompt, and the task is to predict the next item $y^* \in \mathcal{V}$. Each item $v \in \mathcal{V}$ is assigned a \emph{Semantic ID (SID)}~\cite{geng2022recommendation}: an $L$-token code $\mathrm{sid}(v) = (s_1, \ldots, s_L)$ from a learned codebook $\mathcal{S}$ with $|\mathcal{S}| \ll |\mathcal{V}|$, constructed so that semantically related items share prefix tokens.

An LLM with parameters $\theta$ defines the policy
\begin{equation}
    \pi_\theta(y \mid \mathbf{x})
    \;=\; \prod_{\ell=1}^{L} \pi_\theta\bigl(s_\ell \mid \mathbf{x}, s_{<\ell}\bigr),
    \label{eq:policy}
\end{equation}
where the history $\mathbf{x}$ is itself tokenized into SIDs. At inference, candidates are produced by beam search and only sequences mapping to valid items are retained. We use \emph{rollout} for a single sampled sequence and \emph{group} for a set of $K$ rollouts from one prompt.

\subsection{Supervised Fine-Tuning}
\label{sec:prelim:nll}

Given history--target pairs $\mathcal{D} = \{(\mathbf{x}^{(i)}, y^{*(i)})\}_{i=1}^{N}$, the negative log-likelihood (NLL) objective is
\begin{equation}
L_{\text{NLL}}(\theta)
=
- \mathbb{E}_{(\mathbf{x}, y^*) \sim \mathcal{D}}
\left[
\sum_{\ell=1}^{L}
\log \pi_\theta(s_\ell^* \mid \mathbf{x}, s^*_{<\ell})
\right].
\label{eq:nll}
\end{equation}
NLL is stable, anchored to observed behavior, and requires no reward model. It provides the base policy on which RL fine-tuning is applied.

\subsection{Group Relative Policy Optimization}
\label{sec:prelim:grpo}

GRPO~\cite{sheng2024grpo,deepseek2025r1} avoids PPO's value-network overhead by using group-level reward statistics as the baseline. For each prompt $\mathbf{x}$, it draws $K$ rollouts $\{y_1, \ldots, y_K\}$ from $\pi_{\theta_{\text{old}}}$ and scores each: $r_k = \mathrm{RM}(y_k, \mathbf{x})$. The group-relative advantage is
\begin{equation}
    A_k \;=\; \frac{r_k - \bar{r}}{\sigma_r + \epsilon},
    \label{eq:advantage}
\end{equation}
where $\bar{r}$ and $\sigma_r$ are the group mean and standard deviation. With importance ratio $w_k(\theta) = \pi_\theta(y_k \mid \mathbf{x}) / \pi_{\theta_{\text{old}}}(y_k \mid \mathbf{x})$, GRPO optimizes the clipped surrogate
\begin{equation}
\begin{aligned}
    L_{\text{GRPO}}(\theta)
    \;=\; - \mathbb{E}_{\mathbf{x},\, \{y_k\} \sim \pi_{\theta_{\text{old}}}}
        \bigg[ &\frac{1}{K} \sum_{k=1}^{K}
        \min\!\Big(w_k A_k, \\
        &\mathrm{clip}\bigl(w_k,\, 1{-}\epsilon,\, 1{+}\epsilon\bigr) A_k \Big)
        \bigg].
\end{aligned}
\label{eq:grpo}
\end{equation}
This token-level ratio clipping is the mechanism AdaGRPO lifts to the sample level in Section~\ref{sec:method}.
\section{The Reward Model Helps Conditionally: Hard Samples with High RM Discriminability}
\label{sec:analysis}

Before introducing the method, we test empirically whether the reward model provides a useful training signal. The answer is \emph{conditional}---a finding that directly determines what AdaGRPO clips in Section~\ref{sec:method}.

\subsection{Protocol}

For each prompt $\mathbf{x}$ in a held-out set, we run beam search with width $K$ and collect candidates $\mathcal{R} = \{y_1, \dots, y_K\}$. We rank $\mathcal{R}$ by descending beam-search log-probability (\textbf{LLM order}) and by descending reward score (\textbf{RM order}), recording the ground-truth position $\mathrm{idx}_{\mathrm{LLM}}$ and $\mathrm{idx}_{\mathrm{RM}}$ respectively. The signed difference $\Delta = \mathrm{idx}_{\mathrm{LLM}} - \mathrm{idx}_{\mathrm{RM}}$ measures RM \emph{influence}: $\Delta > 0$ means the RM moves the ground truth closer to the top (helpful), $\Delta < 0$ means farther (harmful).

\noindent\textbf{Note.} This analysis uses beam search for reproducibility; GRPO training uses sampling. The distributional gap means the stratified patterns below are indicative rather than a direct measurement of training-time conditions, but they motivate the clip design, which is evaluated end-to-end in Section~\ref{sec:exp}.

\subsection{Aggregate RM Influence Is Near Zero Across All Samples}
\label{sec:analysis_overall}

\begin{table}[t]
  \centering
  \caption{Average ground-truth position under LLM vs.\ RM ordering (\textbf{all samples}). Influence $\Delta = \mathrm{idx}_{\mathrm{LLM}} - \mathrm{idx}_{\mathrm{RM}}$; positive means the RM helps. The net contribution is negligible or negative.}
  \label{tab:rm_overall}
  \begin{tabular}{lccc}
    \toprule
    Beam width $K$ & Avg $\mathrm{idx}_{\mathrm{LLM}}$ & Avg $\mathrm{idx}_{\mathrm{RM}}$ & $\Delta$\\
    \midrule
    50  & 15.79 & 15.31 & $+$0.48 \\
    128 & 35.57 & 35.86 & $-$0.28 \\
    \bottomrule
  \end{tabular}
\end{table}

Table~\ref{tab:rm_overall} shows that the RM's aggregate influence is near zero at $K{=}50$ and turns negative at $K{=}128$. The degradation is consistent with exposure bias: larger beam widths surface out-of-distribution candidates the RM has never been calibrated against. A fixed NLL--GRPO mixture, which trusts the RM uniformly, inherits this near-zero net signal.

\subsection{RM Provides Strong Guidance Only on Hard Samples}
\label{sec:analysis_stratified}

The aggregate picture masks a strong compositional effect. We partition samples by \emph{LLM difficulty}: a sample is \textsc{hard} if its ground truth falls outside the top $\lfloor \tau K \rfloor$ positions under LLM ordering, and \textsc{easy} otherwise---the same partition the clip condition $f_1$ acts on in Section~\ref{sec:method}.

\begin{table}[t]
\small
  \centering
  \caption{Average ground-truth position on \textsc{hard} samples only. The RM provides substantial positive guidance that grows with beam width.}
  \label{tab:rm_hard}
  \begin{tabular}{lcccc}
    \toprule
    $K$ & Avg $\mathrm{idx}_{\mathrm{LLM}}$ & Avg $\mathrm{idx}_{\mathrm{RM}}$ & $\Delta$ & Coverage \\
    \midrule
    50  & 30.07 & 18.66 & $+$11.41 & 42.2\% \\
    128 & 77.09 & 46.33 & $+$30.77 & 33.9\% \\
    \bottomrule
  \end{tabular}
\end{table}

Table~\ref{tab:rm_hard} shows the contrast: the RM now improves ground-truth position by 11.4 ranks at $K{=}50$ and 30.8 at $K{=}128$---the benefit \emph{grows} with candidate-set size, the opposite of the aggregate trend. On the \textsc{easy} partition the influence is near-zero or negative, pulling down the aggregate. The RM is conditionally valuable, and the condition---the policy's own confidence---is observable at training time.

\subsection{RM Discriminability Further Doubles Per-Sample Influence at the Cost of Coverage}
\label{sec:analysis_f2}

Difficulty is necessary but not sufficient. Among \textsc{hard} samples, cases remain where the RM is unreliable---e.g., when the ground truth lies outside its training distribution. We add a second condition mirroring $f_2$ of Section~\ref{sec:method}: retain only \textsc{hard} samples where the RM ranks the ground truth in the top $\lfloor \tau K \rfloor$ \emph{and} every in-batch negative falls below $\lfloor \rho K \rfloor$, so the RM exhibits clear separation between relevant and irrelevant items.

\begin{table}[t]
  \centering
  \caption{Influence after jointly applying both conditions. Per-sample influence roughly doubles, but coverage drops, exposing the precision--coverage trade-off the clip navigates.}
  \label{tab:rm_f1f2}
  \begin{tabular}{lcc}
    \toprule
    $K$ & $\Delta$ & Coverage \\
    \midrule
    50  & $+$23.24 & 11.6\% \\
    128 & $+$59.93 & 13.4\% \\
    \bottomrule
  \end{tabular}
\end{table}

Table~\ref{tab:rm_f1f2} shows that the discriminability condition roughly doubles per-sample influence (nearly 60 ranks at $K{=}128$), but only 12--13\% of samples survive both conditions. The clip resolves this trade-off: it spends RL updates on the high-influence minority and falls back to supervision on the rest.
\section{Method: AdaGRPO}
\label{sec:method}
\begin{figure*}[t]
    \centering
    \includegraphics[width=0.95\textwidth]{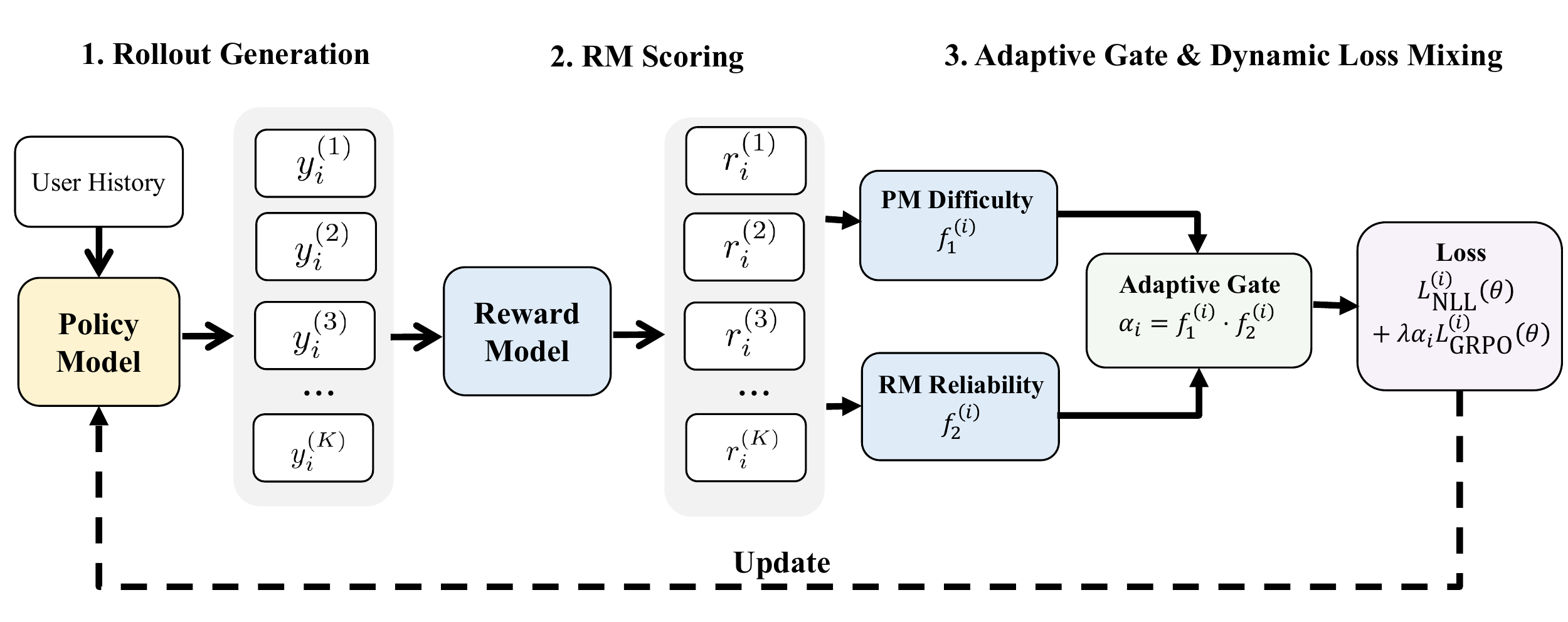}
    \caption{
    Overview of AdaGRPO. Two rank-based diagnostics---one probing policy-side difficulty ($f_1^{(i)}$), the other probing reward-model discriminability ($f_2^{(i)}$)---are evaluated on each training instance and combined into a binary \emph{sample-level clip}. The clip decides, at instance granularity, whether the GRPO loss is admitted into the update or clipped to zero.
    }
    \label{fig:method}
\end{figure*}

AdaGRPO targets the regime in which the reward model supplies informative gradients on only a \emph{proper subset} of training instances. Rather than trusting the reward signal globally, we retain the supervised negative log-likelihood as a stationary anchor and admit the policy-gradient signal through a \emph{sample-level clip}: on each instance the GRPO loss is either passed through intact or clipped to zero, with the clip decision driven entirely by diagnostics on the instance's own rollout group. The per-instance objective is

\begin{equation}
    L_i(\theta)
    \;=\;
    L_{\mathrm{NLL}}^{(i)}(\theta)
    \;+\;
    \lambda\;\alpha_i\;
    L_{\mathrm{GRPO}}^{(i)}(\theta),
    \qquad
    \alpha_i \in \{0,1\},
    \label{eq:hybrid}
\end{equation}

where \(\lambda > 0\) is a global scaling coefficient and \(\alpha_i\) is the \emph{clip coefficient} for instance~\(i\), determined by two rank-based diagnostics on the rollout group. The coefficient is \emph{detached}: no gradient flows through its computation, so clipping reshapes the optimization landscape without introducing a biased surrogate gradient. Setting \(\alpha_i=0\) clips the GRPO term and leaves a pure supervised update; setting \(\alpha_i=1\) admits the full GRPO correction on instance~\(i\).

\paragraph{On the non-triviality of difficulty estimation in recommendation.}
The notion of \emph{sample difficulty} in our setting is fundamentally distinct from its counterpart in difficulty-aware RL for reasoning tasks~\cite{ji2025difficulty}. In mathematical or code reasoning, difficulty admits a clean operational proxy---the \emph{pass rate}---because the reward is sparse and binary and because difficulty is an intrinsic property of the problem, largely decoupled from the policy. Recommendation violates both assumptions. First, there is no policy-independent notion of a ``hard'' user: difficulty is entangled with the current parameterization, the time-varying item catalogue, and the user's idiosyncratic interaction history. Second, the reward signal is \emph{dense}: the production ranker assigns a continuous score to every rollout, rendering pass-rate-style proxies undefined.

These structural differences motivate three desiderata for any quantity that gates a sample-level clip in recommendation: (i)~computability from rollout statistics alone, without auxiliary models or heuristics; (ii)~independence from sparse correctness signals; and (iii)~meaningfulness even when all rollouts receive non-trivial reward scores. The rank-based diagnostics introduced below satisfy all three, and---being read off the same rollout group already materialized for GRPO---add no sampling cost to the clip.

\paragraph{Rank-threshold hyperparameters.}
Both diagnostics convert the rollout group size into rank thresholds through two scalar hyperparameters: a \emph{prominence fraction} \(\tau \in (0,1)\) and a \emph{suppression fraction} \(\rho \in (0,1)\), with \(\tau < \rho\). The prominence threshold \(\lfloor \tau K \rfloor\) marks the top \(\tau\)-fraction of a sorted diagnostic pool, and the suppression threshold \(\lfloor \rho K \rfloor\) marks its complementary tail. Setting \(\tau = 1/3\) recovers a top-tertile threshold \(\lfloor K/3 \rfloor\) and \(\rho = 0.9\) recovers a bottom-decile threshold \(\lfloor 0.9\,K \rfloor\); these defaults are used throughout unless stated otherwise and are empirically stable across tasks. Expressing both thresholds as fractions of \(K\) makes the clip scale-free in the number of rollouts.

\paragraph{Policy-side diagnostic: difficulty.}
For each prompt \(\mathbf{x}_i\), we draw \(K\) rollout candidates
\(\mathcal{R}_i=\{y_i^{(1)},\ldots,y_i^{(K)}\}\)
from the reference policy \(\pi_{\theta_{\mathrm{old}}}\). Let \(y_i^{\star}\) denote the ground-truth target from the supervised dataset. We evaluate \(y_i^{\star}\) under \(\pi_{\theta_{\mathrm{old}}}\) via teacher forcing and compute its length-normalized log-likelihood. Inserting this score into the set of log-likelihoods already computed for the \(K\) rollouts and sorting in descending order yields the \textbf{policy-side diagnostic pool}:

\begin{equation}
    \mathcal{P}_i^{\,\pi}
    \;=\;
    \operatorname{Sort}_{\downarrow}
    \!\Big(
        \big\{\,
            \ell_{\pi}(y) : y \in
            \mathcal{R}_i \cup \{y_i^{\star}\}
        \,\big\}
    \Big),
\end{equation}

where \(\ell_{\pi}(y)\) is the length-normalized log-probability under \(\pi_{\theta_{\mathrm{old}}}\). Let \(\operatorname{rk}_{\pi}(y_i^{\star})\) denote the 1-indexed rank of \(y_i^{\star}\) in \(\mathcal{P}_i^{\,\pi}\). The \textbf{difficulty diagnostic} \(f_1\) is

\begin{equation}
    f_1^{(i)}
    \;=\;
    \mathbbm{1}\!\Big[\,
        \operatorname{rk}_{\pi}(y_i^{\star})
        \;>\;
        \lfloor \tau K \rfloor
    \,\Big].
    \label{eq:c1}
\end{equation}

It fires when the ground truth falls outside the top \(\tau\)-fraction of the policy's own ranking, certifying that the policy does not already assign high likelihood to the correct target. In this regime the supervised gradient alone provides limited corrective signal, so the GRPO term is worth admitting. Conversely, when the ground truth already resides in the top \(\tau\)-fraction the instance is well served by \(L_{\mathrm{NLL}}\) and the GRPO term is redundant---a clear candidate for clipping.

\paragraph{Reward-side diagnostic: reliability.}
Difficulty alone is not a sufficient license to admit the GRPO term: a difficult instance may be precisely one on which the reward model lacks discriminative fidelity, in which case the policy-gradient signal is actively misleading and must be clipped. We therefore add a second diagnostic that probes whether the reward model can separate the ground truth from contextually irrelevant distractors.

For each instance~\(i\), we construct a negative set
\(\mathcal{Z}_i=\{z_i^{(1)},\ldots,z_i^{(M)}\}\)
by uniformly sampling \(M\) rollouts from \emph{other} instances in the same mini-batch. In our experiments we set \(M=5\) throughout; see Section~\ref{sec:discussion} for a discussion of sensitivity to this choice. These serve as contrastive probes: plausible model generations that are contextually irrelevant to \(\mathbf{x}_i\). We score \(y_i^{\star}\) and every element of \(\mathcal{Z}_i\) under the reward model, insert the resulting scores into the rewards already computed for \(\mathcal{R}_i\), and sort in descending order to form the \textbf{reward-side diagnostic pool}:

\begin{equation}
    \mathcal{P}_i^{\,r}
    \;=\;
    \operatorname{Sort}_{\downarrow}
    \!\Big(
        \big\{\,
            r(y \mid \mathbf{x}_i) : y \in
            \mathcal{R}_i \cup \{y_i^{\star}\} \cup \mathcal{Z}_i
        \,\big\}
    \Big),
\end{equation}

where \(r(\cdot \mid \mathbf{x}_i)\) is the reward score conditioned on prompt~\(\mathbf{x}_i\). The \textbf{reliability diagnostic} \(f_2\) is the conjunction of two rank constraints---\emph{ground-truth prominence} and \emph{distractor suppression}:

\begin{equation}
    f_2^{(i)}
    \;=\;
    \mathbbm{1}\!\Big[\,
        \operatorname{rk}_{r}(y_i^{\star})
        \;\le\;
        \lfloor \tau K \rfloor
    \,\Big]
    \;\cdot\;
    \mathbbm{1}\!\bigg[\,
        \min_{m \in [M]}
        \operatorname{rk}_{r}(z_i^{(m)})
        \;>\;
        \lfloor \rho K \rfloor
    \,\bigg].
    \label{eq:c2}
\end{equation}

The first factor (ground-truth prominence) requires the ground truth to reside in the top \(\tau\)-fraction of the reward ranking; the second factor (distractor suppression) requires \emph{every} distractor to fall into the bottom \((1-\rho)\)-fraction. Together they certify clean \emph{rank separation} between contextually relevant and irrelevant outputs. If either sub-condition is violated, the reward model is locally uncalibrated on this instance and \(f_2^{(i)}=0\), so the instance is clipped regardless of its difficulty.

\paragraph{The sample-level clip.}
The clip coefficient is the conjunction of both diagnostics:

\begin{equation}
    \alpha_i
    \;=\;
    f_1^{(i)}
    \;\cdot\;
    f_2^{(i)}.
    \label{eq:alpha}
\end{equation}

The GRPO loss survives the clip if and only if both diagnostics evaluate to unity: the instance must be simultaneously \emph{difficult} for the current policy \emph{and} backed by a \emph{locally reliable} reward signal. Every other instance is clipped to zero and contributes only its supervised update. The two diagnostics thus carve out a complementary failure taxonomy---redundant supervision on one side, unreliable reward on the other---and clip both away, admitting policy-gradient updates only on the instances that survive both tests.

\paragraph{A sample-domain trust region.}
This rule extends the clipping principle of PPO~\cite{schulman2017proximal} from the \emph{ratio domain} to the \emph{sample domain}. PPO clips the importance-sampling ratio to bound the magnitude of each per-token update; AdaGRPO clips entire per-sample losses to exclude instances on which the policy-gradient signal is redundant or unreliable. Both define a trust region by zeroing out updates that fall outside a certified-safe set, but AdaGRPO's trust region is defined over training instances rather than over policy-ratio excursions---a clip on \emph{which} samples may speak, rather than on \emph{how far} each may move.

\paragraph{Design properties.}
The clip is \textbf{hyperparameter-lean}: its only free quantities are the rank-threshold fractions \(\tau\) and \(\rho\), both defined relative to the rollout group size~\(K\) and empirically stable across tasks. It is \textbf{interpretable by construction}: each clip decision is binary, so every instance is unambiguously assigned to the GRPO-active set or its complement, and the two diagnostics expose \emph{why} an instance was clipped, easing diagnosis during training. And it makes \textbf{no global assumption} on reward-model quality: the clip admits the GRPO loss only where both diagnostics locally certify the reward signal as informative, acting as a conservative, instance-wise admission rule for policy-gradient updates.
\section{Experiments}
\label{sec:exp}

\subsection{Experimental Setup}

\paragraph{Dataset.}
We evaluate on proprietary interaction logs from a large-scale e-commerce platform. The training set consists of approximately 175K user--item interaction sequences drawn from a one-week window, and evaluation is conducted on a held-out test set from the two subsequent weeks. This training scale is intentionally smaller than the corpora used for SFT on the same platform, which are on the order of \(10^8\) sequences. Two practical considerations motivate the choice. First, e-commerce traffic is strongly non-stationary: in preliminary studies, expanding the training window to multiple months consistently degraded held-out HR, an outcome consistent with concept drift in user intent and catalogue composition over longer horizons. Second, naive scaling of the RL training set under the standard GRPO recipe triggered reward hacking---HR@10 dropped sharply as the policy increasingly exploited reward-model artefacts---and a robust remedy is still under development. Restricting the training window to one week is therefore a conservative operating point under which all baselines and AdaGRPO can be compared on equal footing. We view scaling AdaGRPO to longer training horizons as an open problem and discuss it as a limitation in Section~\ref{sec:discussion}.

\paragraph{Base model.}
The base policy is a decoder-only LLM supervised-fine-tuned on user--item interaction sequences. Items are represented as semantic token IDs derived from a hierarchical product taxonomy. Unless otherwise stated, each rollout group uses $K{=}50$ sampled candidates, and $M{=}5$ in-batch negatives are drawn for the reward-side diagnostic (Section~\ref{sec:method}).

\paragraph{Reward model.}
We use the CTR head of a production encoder--decoder Transformer ranker~\cite{zhao2019recommending,chen2019behavior} trained on logged click and purchase signals via multi-task learning with a Mixture-of-Experts output layer (AUC $\approx$ 0.76) as an informative but imperfect reward signal.

\paragraph{Baselines.}
We compare against four training variants. \textbf{Base} denotes the SFT-only model. \textbf{GRPO} applies standard policy-gradient training using reward-model scores. \textbf{GRPO+NLL} uses a hybrid objective with a constant mixing coefficient $\lambda$ tuned on a validation split. \textbf{AdaGRPO} denotes our clipped objective, evaluated as \textbf{w.\,$f_1$} with the difficulty condition only and as \textbf{w.\,$f_1$\,\&\,$f_2$} with both the difficulty and reward-discriminability conditions.

\paragraph{Metrics.}
For offline evaluation, we report HR@$k$, ClkRwd@$k$, OrdRwd@$k$ for $k\!\in\!\{1,10,50\}$, together with hallucination rate. HR measures agreement with the held-out next-item target, while ClkRwd and OrdRwd measure reward-model preference under click and order objectives. For online evaluation, we report effective IPV, UCTR, dwell time, and the number of exposed and clicked third-level categories as diversity-related diagnostics.

\subsection{Offline Results}

\begin{figure*}[h]
  \centering
  \includegraphics[width=\linewidth]{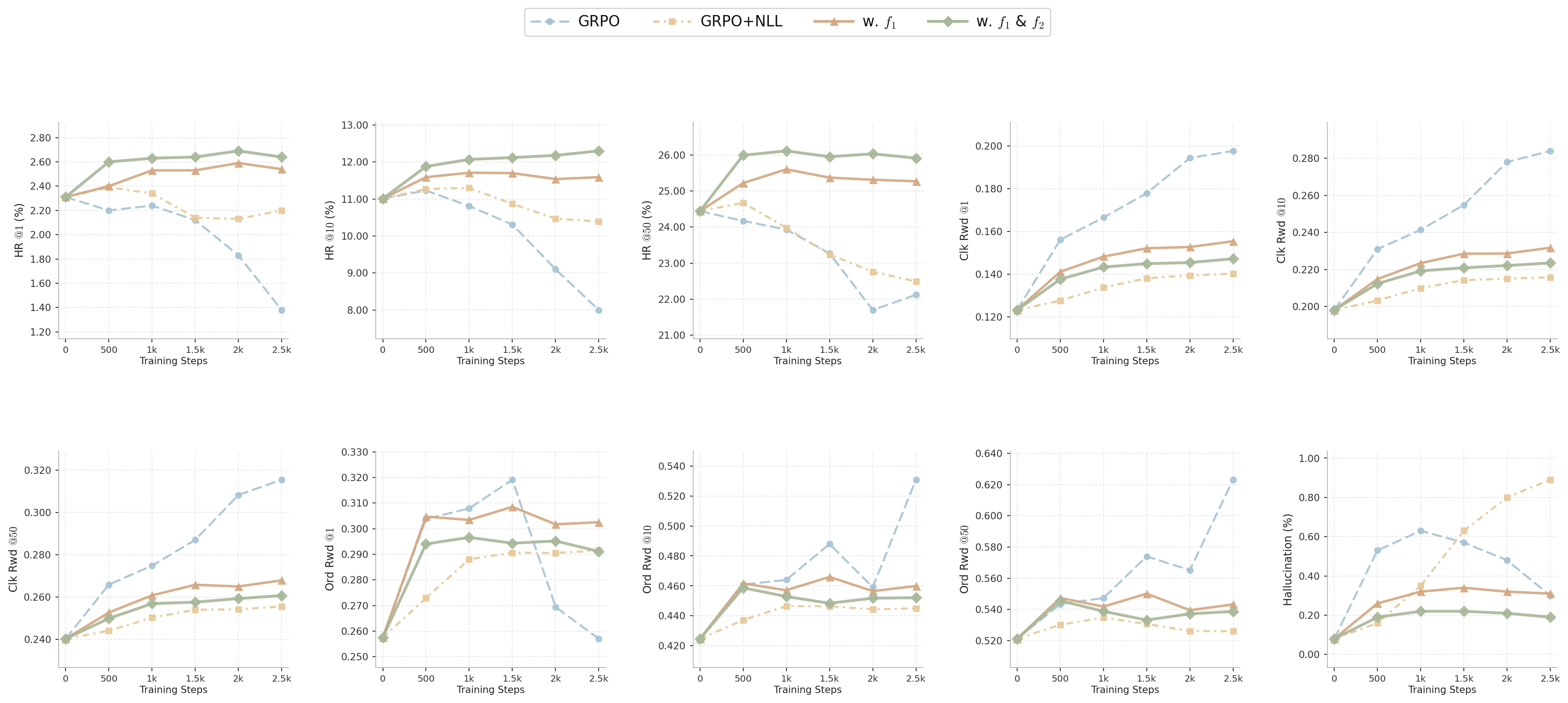}
  \caption{Offline training dynamics over 2{,}500 steps. Standard GRPO increases reward-model scores but is accompanied by a decline in HR@10 and an increase in hallucination rate, suggesting over-optimization to the reward model. GRPO+NLL with a fixed mixing coefficient reduces part of this degradation but does not eliminate the late-stage increase in hallucination. AdaGRPO reaches its best checkpoint at an intermediate training step, improving HR@10 from the Base model's 11.01\% to 12.18\% while keeping hallucination at or below 0.22\%. At the final checkpoint, AdaGRPO remains stable and achieves the best retrieval accuracy among the final models, as detailed in Table~\ref{tab:offline_final}. Adding the reward-discriminability condition $f_2$ further reduces variance and improves the HR--reward--hallucination trade-off.}
  \label{fig:main_results}
\end{figure*}

\begin{table*}[h]
  \caption{Offline performance evaluated at the \textbf{final training checkpoint}. Best intermediate checkpoints are discussed in Figure~\ref{fig:main_results} and the surrounding text; this table reports the final state to highlight training stability. Standard GRPO eventually collapses, trading target retrieval and generation validity for reward maximization, whereas AdaGRPO maintains a stable and superior final-checkpoint trade-off across retrieval accuracy, reward improvement, and hallucination control. The two AdaGRPO rows correspond to the two variants shown in Figure~\ref{fig:main_results}: difficulty-only clipping (w.\,$f_1$) and the full clip with both conditions (w.\,$f_1$\,\&\,$f_2$).}
  \label{tab:offline_final}
  \centering
  \resizebox{\linewidth}{!}{
  \begin{tabular}{lcccccccccc}
    \toprule
    \multirow{2}{*}{Method} & \multicolumn{3}{c}{HR (\%)} & \multicolumn{3}{c}{ClkRwd} & \multicolumn{3}{c}{OrdRwd} & \multirow{2}{*}{Hallucination (\%)} \\
    \cmidrule(lr){2-4} \cmidrule(lr){5-7} \cmidrule(lr){8-10}
    & @1 & @10 & @50 & @1 & @10 & @50 & @1 & @10 & @50 & \\
    \midrule
    Base             & 2.31 & 11.01 & 24.44 & 0.1232 & 0.1981 & 0.2403 & 0.2575 & 0.4247 & 0.5212 & \textbf{0.08} \\
    GRPO             & 2.20 & 10.39 & 22.49 & 0.1402 & 0.2158 & 0.2556 & 0.2915 & 0.4450 & 0.5260 & 0.89 \\
    GRPO+NLL         & 2.30 & 11.06 & 24.14 & \textbf{0.1629} & \textbf{0.2391} & \textbf{0.2747} & \textbf{0.2976} & 0.4479 & 0.5280 & 0.59 \\
    \midrule
    AdaGRPO w.\,$f_1$            & 2.40 & 11.48 & 25.12 & 0.1521 & 0.2345 & 0.2710 & 0.2958 & 0.4589 & 0.5451 & 0.31 \\
    AdaGRPO w.\,$f_1$\,\&\,$f_2$ & \textbf{2.46} & \textbf{11.63} & \textbf{25.43} & 0.1508 & 0.2331 & 0.2698 & 0.2950 & \textbf{0.4617} & \textbf{0.5487} & 0.27 \\
    \bottomrule
  \end{tabular}
  }
\end{table*}

Figure~\ref{fig:main_results} summarizes the offline training dynamics, including both the best intermediate checkpoints and the final training state. Table~\ref{tab:offline_final} reports the exact performance at the final training checkpoint, so the table should be interpreted as a stability comparison rather than as a best-checkpoint selection. The results support three observations.

\paragraph{Reward-only GRPO improves RM scores but suffers from late-stage collapse.}
While GRPO steadily increases ClkRwd and OrdRwd metrics during training, the policy eventually collapses under over-optimization. As shown in Table~\ref{tab:offline_final}, at the final checkpoint, HR@10 and HR@50 significantly degrade from the Base model's 11.01\% and 24.44\% down to 10.39\% and 22.49\%, respectively. Simultaneously, the hallucination rate spikes severely to 0.89\%. This pattern indicates a fundamental misalignment: optimizing solely for the reward model actively destroys the policy's structural knowledge for held-out target retrieval.

\paragraph{A fixed hybrid objective is more stable but remains insufficient.}
GRPO+NLL slows the decline in retrieval metrics relative to GRPO-only training, roughly maintaining the base HR@10 (11.06\% at final step). However, the hallucination rate still drifts upwards, reaching 0.59\% at the end of training. This confirms the central premise of our clip: a single global mixing coefficient cannot account for sample-level variation in reward-model reliability, so it inevitably leaks noisy reward signal into the gradient on the instances where the RM is uninformative.

\paragraph{AdaGRPO ensures stability and improves the HR--reward--hallucination trade-off.}
In contrast to the baselines, AdaGRPO safely lifts performance without collapsing. Along the training trajectory, AdaGRPO (w.\,$f_1$\,\&\,$f_2$) reaches a best checkpoint with HR@10 of 12.18\%, improving over the Base model's 11.01\%, while keeping hallucination at or below 0.22\%. This best-checkpoint result reflects the peak retrieval--validity trade-off observed in the dynamic curves of Figure~\ref{fig:main_results}. At the final checkpoint, which is the setting reported in Table~\ref{tab:offline_final}, AdaGRPO (w.\,$f_1$\,\&\,$f_2$) still yields the best retrieval accuracy among the final models (HR@10 of 11.63\%, HR@50 of 25.43\%) while keeping hallucination controlled at 0.27\%. Comparing the two AdaGRPO rows, adding the reward-discriminability condition $f_2$ further reduces hallucination (from 0.31\% to 0.27\%) and improves HR, confirming the value of the reward-side diagnostic. This stability comes with a mild and acceptable reduction in the absolute ClkRwd peak relative to the fixed variant, which is consistent with the design intent of the clip: it withholds RL updates in less reliable reward regions to preserve long-term generation quality.

\paragraph{Difficulty-stratified analysis.}
We further partition the test set into five equal-mass bins according to the ground-truth rank under the base model's rollout. Table~\ref{tab:stratified} reports the per-bin HR@10 for GRPO+NLL and AdaGRPO (w.\,$f_1$\,\&\,$f_2$). The HR@10 gain is largest in the intermediate-difficulty region, with a gain of $+0.017$ over GRPO+NLL in the 40--80\% difficulty range. The gain is mildly negative in the easiest bin, $-0.015$, and close to zero in the hardest bin. This pattern is consistent with the intended behavior of the clip: it admits RL updates where the policy is uncertain and the reward is reliable, clips them where the policy is already confident, and clips them again where the rollout set lies too far from the reward model's reliable region. The analysis supports the clipping mechanism, but it should be read as diagnostic evidence rather than as a complete validation of the causal mechanism.

\begin{table}[h]
  \centering
  \caption{Difficulty-stratified HR@10 (\%) at the final checkpoint. Bins are defined by the ground-truth rank percentile under the base model's rollout (0--20\% is easiest, 80--100\% is hardest). $\Delta$ denotes AdaGRPO minus GRPO+NLL.}
  \label{tab:stratified}
  \begin{tabular}{lccc}
    \toprule
    Difficulty bin & GRPO+NLL & AdaGRPO & $\Delta$ \\
    \midrule
    0--20\%  (easiest) & 24.31 & 24.16 & $-$0.015 \\
    20--40\%           & 12.83 & 12.90 & $+$0.007 \\
    40--60\%           &  8.47 &  8.64 & $+$0.017 \\
    60--80\%           &  5.21 &  5.38 & $+$0.017 \\
    80--100\% (hardest)&  2.48 &  2.49 & $+$0.001 \\
    \bottomrule
  \end{tabular}
\end{table}

\subsection{Online A/B Test}

\begin{table}[h]
  \caption{Online A/B results. Each number is reported as lift relative to the contemporaneous production control in its corresponding experiment. GRPO+NLL ran from 01/03 to 15/03, and AdaGRPO ran from 24/03 to 31/03. Because the two experiments were conducted in \emph{different time windows} against \emph{different controls}, the columns should not be compared directly as a head-to-head result; they are presented together for compactness only. Superscript $\ast$ denotes ${p}_{\mathrm{value}}<0.05$.}
  \label{tab:online}
  \centering
  \setlength{\tabcolsep}{8pt}
  \begin{tabular}{lcc}
    \toprule
    Metric & GRPO+NLL & AdaGRPO \\
    \midrule
    Eff.\ IPV      & $+$0.09\%      & $+0.43\%^\ast$ \\
    Strict IPV     & $+$0.14\%      & $+0.35\%^\ast$ \\
    UCTR           & $-$0.09\%      & $+0.27\%^\ast$ \\
    Dwell time     & $+$0.01\%      & $+0.23\%^\ast$ \\
    \midrule
    Exposed cats.\ & $+0.14\%^\ast$ & $+0.25\%^\ast$ \\
    Clicked cats.\ & $+0.16\%^\ast$ & $+0.28\%^\ast$ \\
    \bottomrule
  \end{tabular}
\end{table}

Table~\ref{tab:online} reports online results from two production A/B tests. GRPO+NLL shows small and statistically non-significant changes on the main engagement metrics, despite improving some offline reward metrics. This suggests that higher reward-model scores alone do not necessarily translate into measurable user engagement gains.

AdaGRPO shows statistically significant positive lifts on the reported engagement metrics, including effective IPV, UCTR, and dwell time. It also increases the number of exposed and clicked third-level categories, suggesting that the clip does not collapse the policy toward a narrower set of reward-favored categories in this deployment. These online results are consistent with the offline diagnosis that clipping GRPO updates to locally reliable instances improves the usefulness of reward-model training. However, because the two experiments were conducted in different time windows against different contemporaneous controls, the evidence should be viewed as production support for each method relative to its own control, rather than as a definitive head-to-head causal comparison between GRPO+NLL and AdaGRPO.

Overall, the offline and online results indicate that AdaGRPO can improve the trade-off between reward optimization, target retrieval, and hallucination control in this industrial recommendation setting. The evidence is strongest for the claim that a sample-level clip makes GRPO training more robust to reward-model noise. Broader claims about general deployment robustness require additional A/B tests across traffic slices, user segments, item popularity levels, and catalogue freshness regimes.
\section{Discussion}
\label{sec:discussion}

AdaGRPO's clip is not naive filtering: it acts solely on the GRPO term while the NLL term remains active for all instances, ensuring that even clipped samples continue to anchor the model's recommendation behavior through supervision. The clip operates on rank quantiles relative to $K$ ($\lfloor \tau K \rfloor$, $\lfloor \rho K \rfloor$), so decisions depend on the instance's position within the diagnostic pool rather than absolute score magnitudes, making it a conservative admission rule rather than a proof of reward model correctness. The diagnostics $f_1$ and $f_2$ serve as computable proxies, identifying policy uncertainty and reward-model discriminability, and a surviving clip merely indicates that an instance is suitable for a GRPO update under these diagnostics, not that the reward model is perfectly calibrated or causally aligned. This principle applies in settings where supervised targets are available, reward signals are noisy, and rollout diagnostics can detect locally informative guidance, though domains lacking a ranked candidate set, identifiable ground truth, or meaningful in-batch negatives may require alternative diagnostics. Limitations include the introduction of hyperparameters $\tau$, $\rho$, $\lambda$, and $M$, which require validation-based tuning, sensitivity of $f_2$ to batch composition, reliance on ground-truth items during training, and the fact that online evaluation was conducted in a single production setting with separate A/B test windows, emphasizing the need for extended experiments across user segments, item popularity levels, and catalogue freshness regimes to establish broader robustness.
\section{Conclusion}
\label{sec:conclusion}

We presented AdaGRPO, a sample-level adaptive framework for RL fine-tuning of generative recommenders. By anchoring training on supervised NLL and admitting the GRPO term only when a binary clip certifies both policy-side difficulty and reward-model discriminability, AdaGRPO addresses the core tension of using noisy production rankers as reward sources. Empirically, the RM's aggregate influence is near zero, but it provides strong guidance on the subset of hard samples where it is locally reliable. AdaGRPO exploits this structure: offline, it improves HR@10 from 11.01\% to 12.18\% at the best checkpoint while controlling hallucination, and remains the strongest final-checkpoint model. Online, it delivers statistically significant engagement gains. Our results suggest that the central challenge of RL for generative recommendation is not designing stronger rewards, but knowing when to trust them.

\section{Speaker Bio} \textbf{Yanyan Zou} is an applied scientist in Recommendation Platform at JD.com since 2020, launching cutting-edge AI models into practical productions.
Her research interests primarily lie in the areas of large language model and recommendation, with around 20 papers published in top-tier conferences (e.g., ACL, EMNLP, AAAI). She received her B.Engr. degree in 2015 from Xiamen University, China, as well as her Ph.D. degree from
	Singapore University of Technology and Design
	in 2020. 

\bibliographystyle{ACM-Reference-Format}
\bibliography{main}

\end{document}